\documentclass{article}

\usepackage[english]{babel}

\usepackage[letterpaper,top=2cm,bottom=2cm,left=3cm,right=3cm,marginparwidth=1.75cm]{geometry}

\usepackage{amsmath}
\usepackage{amssymb}
\usepackage{graphicx}
\usepackage{subcaption}  
\usepackage[colorlinks=true, allcolors=blue]{hyperref}
\usepackage{braket}
\usepackage{authblk}
\usepackage{mathtools}
\usepackage{hyperref}
\usepackage{url}
\usepackage{float}
\usepackage[toc,page]{appendix}
\usepackage{titlesec}

\title{HiPPO-KAN: Efficient KAN Model for Time Series Analysis}
\author{SangJong Lee, Jin-Kwang Kim, JunHo Kim, TaeHan Kim, James Lee}
\affil{XaaH Corp\\ \texttt{\{sangjong, jinkwang, demyank, taehankim, jaminyx\}@xaah.xyz}}
\date{}

\begin{document}
\maketitle

\begin{abstract}
In this study, we introduces a parameter-efficient model that outperforms traditional models in time series forecasting, by integrating High-order Polynomial Projection (HiPPO) theory into the Kolmogorov-Arnold network (KAN) framework. This HiPPO-KAN model achieves superior performance on long sequence data without increasing parameter count. Experimental results demonstrate that HiPPO-KAN maintains a constant parameter count while varying window sizes and prediction horizons, in contrast to KAN, whose parameter count increases linearly with window size. Surprisingly, although the HiPPO-KAN model keeps a constant parameter count as increasing window size, it significantly outperforms KAN model at larger window sizes. These results indicate that HiPPO-KAN offers significant parameter efficiency and scalability advantages for time series forecasting. Additionally, we address the lagging problem commonly encountered in time series forecasting models, where predictions fail to promptly capture sudden changes in the data. We achieve this by modifying the loss function to compute the MSE directly on the coefficient vectors in the HiPPO domain. This adjustment effectively resolves the lagging problem, resulting in predictions that closely follow the actual time series data. By incorporating HiPPO theory into KAN, this study showcases an efficient approach for handling long sequences with improved predictive accuracy, offering practical contributions for applications in large-scale time series data. 
\end{abstract}

\section{Introduction}

The purpose of deep learning is to find a well-approximated function, especially when a target function involves non-linearity or high-dimensionality. In the case of Multilayer perceptron (MLP), a representational ability for non-linear functions is guaranteed by the universal approximation theorem \cite{geometric_dl, textbook1}. Recently, Kolomogorov-Arnold network (KAN) has been proposed as a promising alternative to MLP \cite{kan1, kan2}. This model is distinct to MLP in that it learns activations functions rather than weigths of edges. Its representational ability is partially guaranteed by Kolomogorov-Arnold theorem (KAT), while KAN slightly reshapes the theorem by assuming smooth activation functions but with a deeper network so that it can utilize the back-propagation mechanism. It outperforms MLP with a better scaling law, offering new pathway to modeling complex functions.

Numerous models based on MLP, RNN, and LSTM have been developed to conduct time-series analysis, aiming to model complex patterns and non-linearities \cite{MLP_timeseries_forecasting1, MLP_timeseries_forecasting2, RNN_timeseries_forecasting, LSTM_timeseries_forecasting1, LSTM_timeseries_forecasting2}. Also, deep state space model has emerged as a powerful approach for time series forecasting \cite{DSSM_timeseries_forecasting1}. It combines the strengths of traditional state space models with the representation-learning capabilities of deep learning, allowing them to capture complex temporal dynamics more effectively. By integrating probabilistic reasoning and deep learning, the deep state space model offers significant improvements in forecasting time-series data \cite{DSSM_timeseries_forecasting1, DSSM_timeseries_forecasting2, DSSM_timeseries_forecasting3}. While these methods have shown some success in capturing trends and seasonality, they often face challenges in capturing complex patterns and, especially learning long-term dependencies \cite{rnn_vanishing_grad}.

Long-term dependency is particularly important in time series analysis, as many real-world datasets such as those in finance, weather forecasting, and energy consumption, involve patterns that evolve over long periods. Capturing these dependencies enables models to make more accurate predictions by considering not only short-term fluctuations but also broader trends and delayed effects that span across extended windows. To address these issues, A. Gu et al. introduced the HiPPO (High-order Polynomial Projection Operator) theory and the Structured State Space (S4) model \cite{hippo, S4, s4d, hippo-train}, which effectively capture long-range dependencies by performing online function approximation with special initial conditions in the state space transition equation.

In this work, we build upon the HiPPO theory to enhance the capabilities of the Kolmogorov-Arnold Network (KAN) for time series analysis. According to the HiPPO (High-order Polynomial Projection Operator) theory, a special combination of matrices $A$ and $B$ in the transition equation of a state space model enables the mapping of sequential data, such as time series, into a finite-dimensional space expanded by well-defined polynomial bases. This means that time series data can be represented as a coefficient vector whose dimension is independent of the length of the sequence.

\begin{figure}[t]
    \centering
    \includegraphics[width=\textwidth]{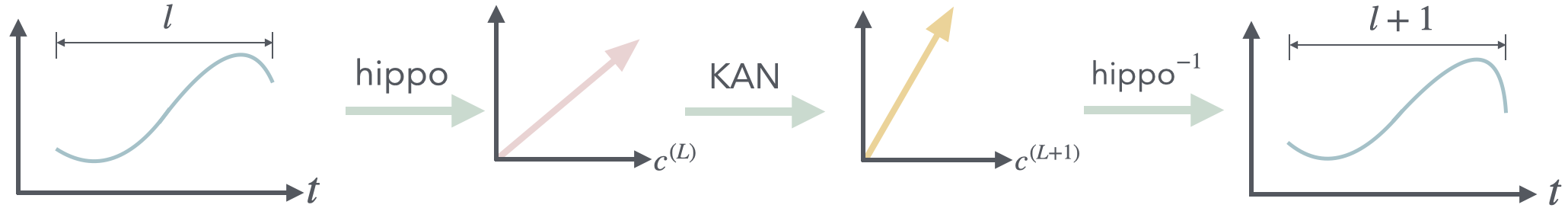}
    \caption{This diagram illustrates the process of encoding time series data using the HiPPO framework, transforming it with the Kolmogorov-Arnold Network (KAN), and decoding it back to the time domain. The initial time series \( l \) is projected into a coefficient vector \( c^{(L)} \) through HiPPO. This vector is then transformed by KAN into \( c^{(L+1)} \), followed by decoding through HiPPO to reconstruct the time series of length \( l+1 \). This setup serves as an auto-encoder where HiPPO and KAN handle encoding, transformation, and decoding, respectively.}
    \label{fig:hippo_kan_diagram}
\end{figure}

Leveraging this property, we can effectively forecast future time-series with smaller parameters. Figure~\ref{fig:hippo_kan_diagram} exhibits a schematic of the HiPPO-KAN process. It first encodes the time series data into a fixed-dimensional coefficient vector using the HiPPO framework. Then it maps this coefficient vector into another vector within the same dimensional space. This mapping is performed using the Kolmogorov-Arnold Network (KAN), which acts as a function approximator in this context. Finally, we decode the transformed coefficient vector back into the time domain using the inverse function provided by the HiPPO framework. This process is analogous to an auto-encoder, where the encoder and decoder are defined by the HiPPO transformations, and the latent space manipulation is handled by KAN. 

Accordingly, our contributions are as follows:
\begin{itemize}
    \item[1.] \textbf{Parameter Efficiency and Scalability in Univariate Time Series Prediction.} We demonstrate that HiPPO-KAN achieves superior parameter efficiency in univariate time-series prediction tasks. The dimension of the coefficient vector remains fixed regardless of the input sequence length, enabling the model to scale to long sequences without increasing the number of parameters. This scalability is crucial for practical applications involving large datasets.
    \item[2.] \textbf{Enhanced Performance Over Traditional KAN in Long-Range Forecasting.} We show that HiPPO-KAN outperforms the traditional KAN as well as other traditional models specialized to handle sequential data (e.g. RNN and LSTM), especially in long-range forecasting scenarios. By effectively capturing long-term dependencies through the HiPPO framework, our model provides more accurate predictions compared to KAN alone.
    \item[3.] \textbf{Novel Integration of HiPPO Theory with KAN.} The use of HiPPO coefficients provides a concise and interpretable state representation of the time series system. When combined with KAN's transparent architecture, this allows for better understanding and interpretability of the model's internal workings.
\end{itemize}

\section{Backgrounds}
\subsection{State Space Model}
State space model can be written as
\begin{align}\label{transition}
    \frac{\mathrm{d}}{\mathrm{d}t}\mathbf{x}(t) &= \mathbf{A}\mathbf{x}(t) + \mathbf{B} \mathbf{u}(t), \\
    \mathbf{y}(t) &= \mathbf{C} \mathbf{x}(t) + \mathbf{D} \mathbf{u}(t), \label{observation}
\end{align}
where $\mathbf{u}(t) \in \mathbb{R}^{l}$ is an input vector, $\mathbf{x}(t) \in \mathbb{R}^N$ is a hidden state vector, and $\mathbf{y}(t) \in \mathbb{R}^{k}$ is an output vector. Eq.(\ref{transition}) describes the state dynamics, showing how the state $\mathbf{x}(t)$ evolves over time based on its current value and the input $\mathbf{u}(t)$. The matrix $\mathbf{A} \in \mathbb{R}^{N \times N}$ defines the influence of the current state on its rate of change, while $\mathbf{B} \in \mathbb{R}^{N \times l}$ defines how the input affects the state dynamics. Eq.(\ref{observation}) represents the output equation, illustrating how the current state and input produce the output $\mathbf{y}(t)$. The matrix $\mathbf{C} \in \mathbb{R}^{k \times N}$ maps the state to the output, and $\mathbf{D} \in \mathbb{R}^{k \times l}$ maps the input directly to the output.

In many cases, especially when implementing skip connections akin to those in deep learning architectures, we can set $\mathbf{D} = 0$. This simplifies the output equation to
\begin{align}
    \mathbf{y}(t) = \mathbf{C} \mathbf{x}(t).
\end{align}
By doing so, the output depends solely on the internal state, allowing the model to focus on the learned representations within $\mathbf{x}(t)$ without direct influence from the immediate input $\mathbf{u}(t)$. Gu et al. showed that when the system is linear time-invariant (LTI), the SSM reduces to a sequence-to-sequence mapping by defining a convolution mapping
\begin{align}
    K(t) = \mathbf{C}e^{t\mathbf{A}} \mathbf{B}, \quad \mathbf{y}(t) = (\mathbf{K} * \mathbf{u})(t).
\end{align}
Gu et al.\cite{hippo} also showed that, by selecting specific initial conditions for the parameters $(\mathbf{A},\mathbf{B})$, $e^{t\mathbf{A}}\mathbf{B}$ becomes a vector of $N$ basis functions. This result enables the state-space model to perform online function approximation using the HiPPO theory.

\subsection{HiPPO Theory}
The memorization process can be considered as a symmetry breaking of fully symmetric states. The fully symmetric state represents a system with maximum entropy, where all configurations are equally probable. By introducing the information we wish to memorize as an external condition, we break this symmetry, allowing the system to settle into a specific state that encodes the memory.

In the context of continuous time series, this approach was exemplified by the Legendre Memory Unit (LMU) \cite{LMU}. The LMU employs continuous orthogonal functions—specifically, Legendre polynomials—to maintain a compressed representation of the entire history of input data. Building upon these principles, Gu et al. \cite{hippo} connected memorization to state-space models with a strong theoretical foundation. Specifically, they demonstrated that a special initialization of the transition equation in the state-space model enables closed-form function approximation, effectively capturing long-term dependencies in sequential data. HiPPO treats memorization as an online function approximation.

Suppose we have a univariate time series function:

\begin{align}
f: \mathbb{R}_{\geq 0} \rightarrow \mathbb{R}, \quad t \mapsto f(t).
\end{align}
Since we are considering online function approximation, we define:
\begin{align}\label{Hilbert space}
    x_n(t) = \int_0^t \mathrm{d}s\,\omega(t,s) p_n(t, s)u(s), \quad \left<p_n,\, p_m\right>_\omega \equiv \int_0^t \mathrm{d}s\, \omega(t,s) p_n(t,s)p_m(t,s) = \delta_{n,m},
\end{align}
which states that for every fixed $t$, the function $p_n$ belong to a Hilbert space $\mathcal{H}$ and form an orthonormal basis with respect to the measure $\omega(t,s)$. Rewriting Eq.(\ref{Hilbert space}), we have:
\begin{align}
    x_n(t) = \int_0^t \mathrm{d}s \omega(t,s)\, p_n(t,s) u(s) = \left<u,\,p_n(t)\right>_\omega,
\end{align}
which indicates that the state vector $\mathbf{x}(t) = [x_1(t),x_2(t),\ldots,x_N(t)]^T$ represents the projection of $u(s)$ for $s\leq t$ onto an orthonormal basis with respect to weighted inner product defined by $\omega(t,s)$.

If we assume completeness, we have:
\begin{align}
    u(s) = \lim_{N\rightarrow \infty} \sum_{n=1}^N x_n(t) p_n(t, \,s)
\end{align}
for all $s \leq t$ due to the completeness of the basis function. Since we are dealing with a finite $N$, by choosing an appropriate cutoff, we obtain an approximate representation of the function $u(s)$. Gu et al. \cite{hippo-train} defined this problem as online function approximation in the HiPPO theory.

\begin{figure}[h]
    \centering
    \begin{subfigure}[b]{0.48\textwidth}
        \centering
        \includegraphics[width=\textwidth]{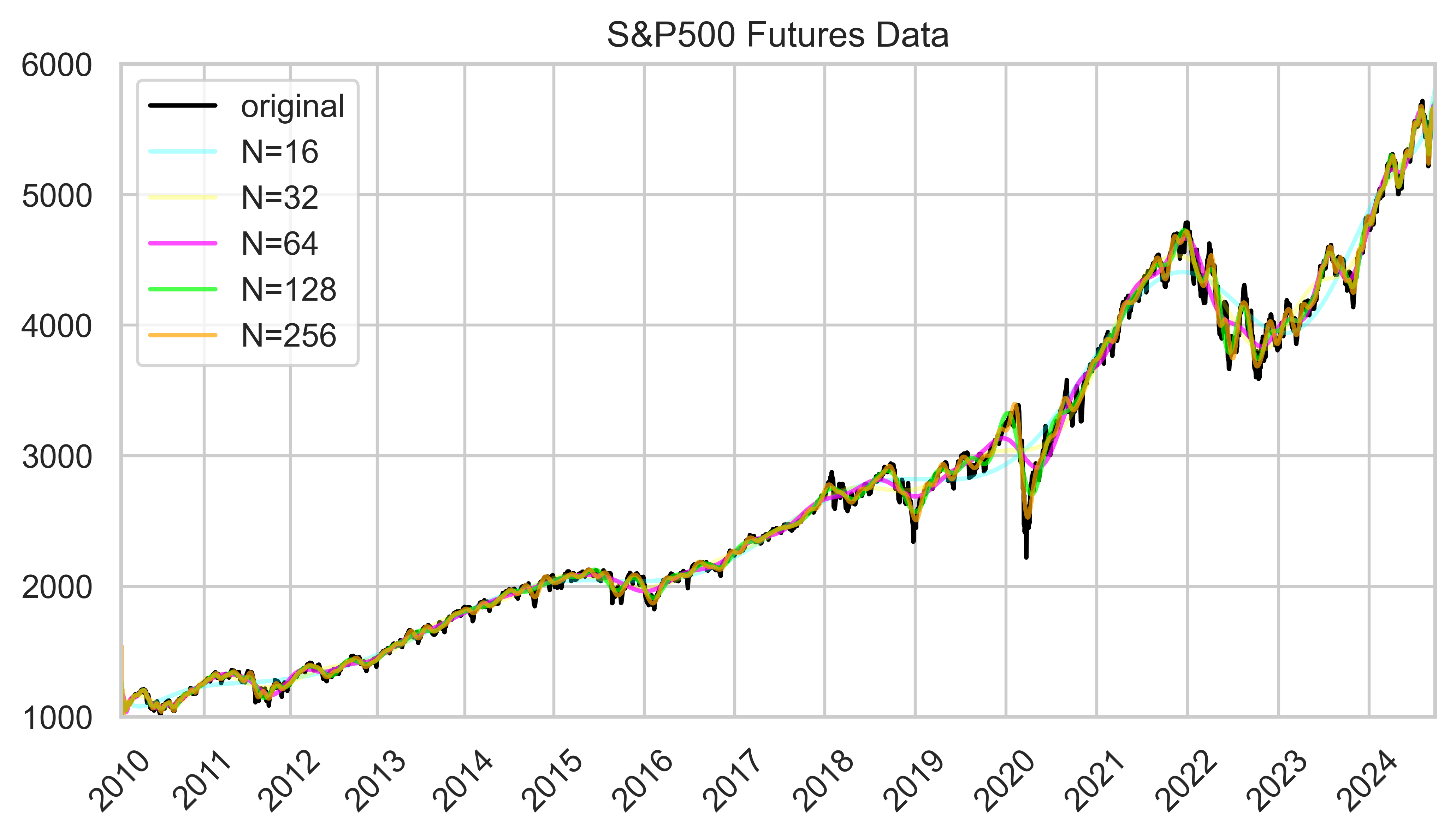}
        \caption{Comparison of S\&P500 data with approximated data using HiPPO for different state space dimensions.}
        \label{fig:approx1}
    \end{subfigure}
    \hfill
    \begin{subfigure}[b]{0.48\textwidth}
        \centering
        \includegraphics[width=\textwidth]{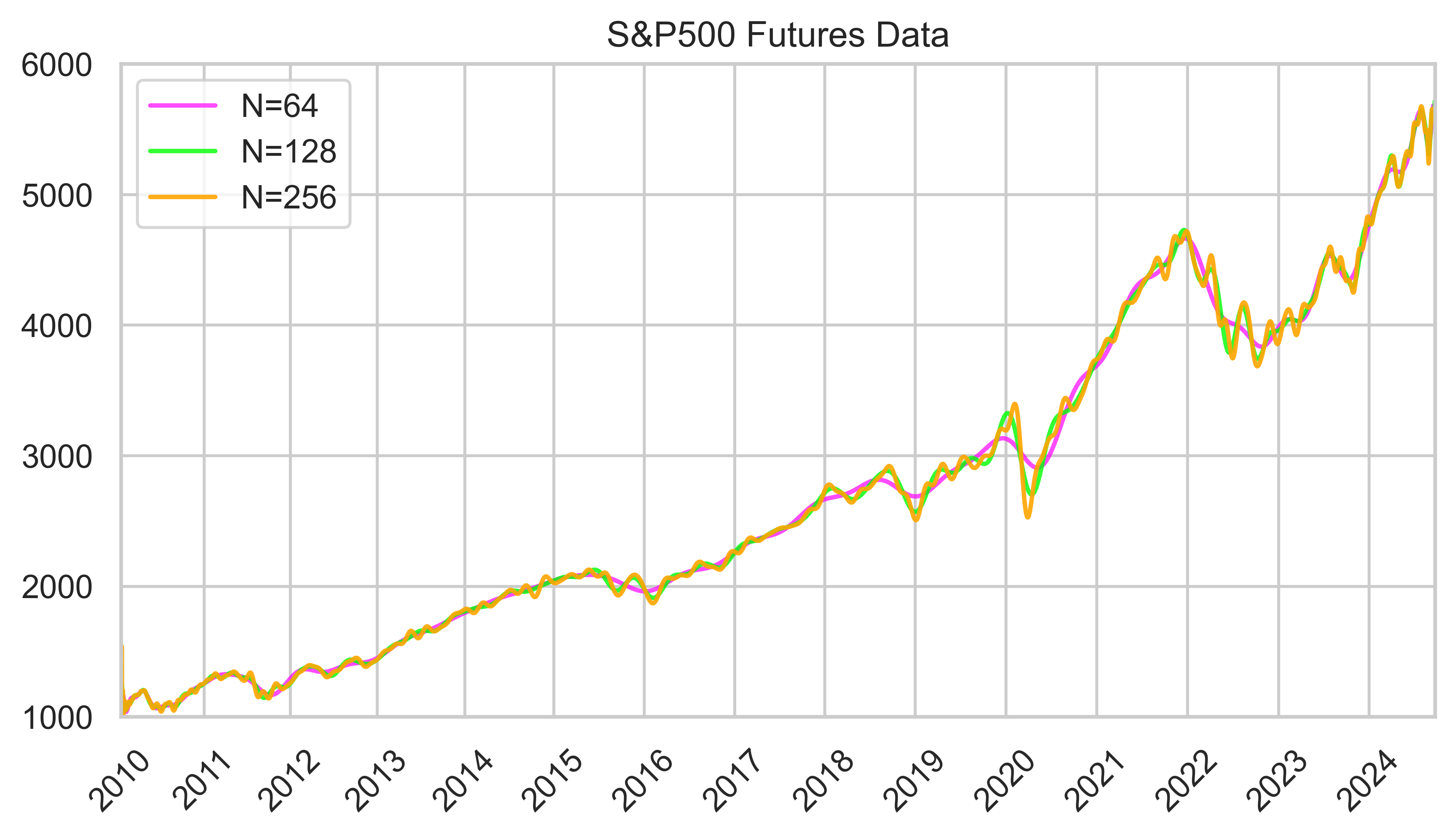}
        \caption{HiPPO approximation of S\&P500 data for state space dimensions $N=64, 128, 256$, showing progressively closer fits to the original data as $N$ increases.}
        \label{fig:approx2}
    \end{subfigure}
    \caption{HiPPO was applied to the S\&P 500 data. The state space dimension used were $N=16,32,64,128,256$, and as $N$ increases, the approximation becomes increasingly closer to the original function, reflecting a higher fidelity representation of the underlying dynamics.}
    \label{fig:approx_ex}
\end{figure}

From a physical standpoint, this is analogous to a multipole expansion, where each term has a specific physical interpretation. In the case of a nonlinear function that takes the coefficients of a multipole expansion as inputs, each coefficient corresponds to a node within the function. Ideally, during the process of learning this nonlinear function, deriving a closed-form solution or understanding how each node operates would greatly aid in physical interpretation. This understanding can provide significant insights into the underlying physics and how the model represents the system. To further enhance this interpretability, we utilized KAN to model the mapping from the coefficients of sequential data of length $l$ to sequential data of length $l+1$.

\subsection{KAN}
\subsubsection{Kolmogorov-Arnold Theorem}
The Kolmogorov-Arnold Representation Theorem states that any continuous multivariate function $f$ defined on a bounded domain $I^n$, where $n$ is the number of variables and $I=[0,1]$, can be expressed as a finite sum of compositions of continuous univariate functions and addition. Specifically, for a smooth function $f$:
\begin{align}
    f:I^n \rightarrow \mathbb{R}, \quad \mathbf{x} \in I^n \mapsto f(x_1,\cdots, x_n) = \sum_{q=1}^{2n+1} \Phi_q \left( \sum_{p=1}^n \phi_{q,p}(x_p) \right),
\end{align}
where each $\phi_{q,p}: I \rightarrow \mathbb{R}$ and $\Phi_q: \mathbb{R} \rightarrow \mathbb{R}$ are continuous univariate functions. This theorem reveals that any multivariate continuous function can be constructed using only univariate continuous functions and addition, significantly simplifying their analysis and approximation. This decomposition reduces the complexity inherent in multivariate functions, making them more tractable for approximation methods. 

\subsubsection{Kolmogorov-Arnold Network}
Building upon the Kolmogorov-Arnold representation theorem, the Kolmogorov-Arnold Network (KAN) is designed to explicitly parametrize this representation for practical function approximation in neural networks \cite{kan1, kan2}. Since we have decomposed the multivariate function into univariate functions, the problem reduces to parametrizing these univariate functions. To achieve this, we can use B-splines due to their flexibility and smoothness properties, which are advantageous for interpolations.
From the perspective of generalizing the Kolmogorov–Arnold (KA) representation theorem and extending it to deeper networks, the network architecture can be expressed as follows:
\begin{align}
    [n_0,n_1,\cdots,n_L],
\end{align}
where $n_l$ is the number of nodes in the $l$-th layer. The pre-activation values are given by:
\begin{align}
    x_{l+1,j} = \sum_{i=1}^{n_l} \phi_{l,j,i}(x_{l,i}), \quad l = 0,\ldots, L-1; \; j = 1, \ldots, n_{l+1}
\end{align}
where $\phi_{l,j,i}$ are the univariate functions with learnable parameters in the $l$-th layer. 

In practice, the univariate functions $\phi_{l,j,i}$ in KAN are parametrized using B-splines to capture complex nonlinearities while maintaining smoothness and flexibility. To enhance the representational capacity of the network and facilitate efficient training, KAN employs residual activation functions that combine a basis function with a spline function. Specifically, the activation function at each node is defined as
\begin{align}
    \phi(x) = w_b \, b(x) + w_s \, \text{spline}(x)
\end{align}
where $b(x)$ is a predefined basis function, $w_b$ and $w_s$ are learnable weights, and $\text{spline}(x)$ is a spline function constructed from B-spline basis functions. The basis function $b(x)$ is typically chosen as the SiLU (Sigmoid Linear Unit) activation function due to its smoothness and nonlinearity.

The overall network function is then:
\begin{align}
    \text{KAN}(\mathbf{x}) = (\Phi_{L-1}\circ\Phi_{L-2}\circ \cdots \circ \Phi_0)(\mathbf{x}).
\end{align}
In this expression, $\Phi_l$ represents the vector of univariate functions at layer $l$, and the composition of these functions across layers forms the basis of KAN's ability to approximate multivariate functions. 

In the context our work, we extend KAN by integrating with the HiPPO framework to efficiently handle time series data. This integration allows us to leverage KAN's function approximation capabilities while benefiting from HiPPO's ability to represent sequential data in a fixed-dimensional space. 

\subsection{Time-series forecasting using KAN}
Since its introduction, KAN have been proved to be a powerful tool for time-series forecasting due to their effective approximation capabilities and training efficiency. It has been shown that KAN models outperform MLP models in time-series forecasting, both in terms of accuracy and computational efficiency \cite{vaca_kan_time,xu_kan_time}. Furthermore, when KAN layers are incorporated within recurrent neural networks (RNNs) and transformer architectures, they excel in multi-horizon forecasting tasks with reduced overfitting issues \cite{genet_tkan,genet_tkat}. 

While these approaches validate the effectiveness of KAN models in time-series prediction and outperforms traditional models specialized in sequential data (e.g., RNN and GRU), they involve integrating KAN into complex architectures, which can increases model complexity and computational demands. In this study, however, we propose an alternative methodology that combines KAN models with HiPPO transformation. By integrating KAN with the HiPPO transformation, we construct a simpler model architecture that retains high predictive performance without relying on complex recurrent or transformer structures.

\section{HiPPO-KAN}
Building upon the HiPPO framework, we consider a univariate time series $u_{1:L}\in\mathbb{R}^L$. The HiPPO transformation maps this time series into a coefficient vector $\mathbf{c}^{(L)} \in \mathbb{R}^N$ via the mapping
\begin{align}
    \text{hippo}_L: \mathbb{R}^L \rightarrow \mathbb{R}^N, \quad u_{1:L} \mapsto \mathbf{c}^{(L)} = \text{hippo}_L(u_{1:L}),
\end{align}
where $N$ is the dimension of the hidden state. In our proposed method, the KAN is utilized to model the mapping between coefficient vectors corresponding to time series of length $L$ and $L+1$. Specifically, KAN transforms the coefficient vector $\mathbf{c}^{(L)}$ into a new coefficient vector $\mathbf{c}^{(L+1)}$:
\begin{align}
    \text{KAN}: \mathbb{R}^N \rightarrow \mathbb{R}^N, \quad \mathbf{c}^{(L)} \mapsto \mathbf{c}^{(L+1)} = \text{KAN}(\mathbf{c}^{(L)}).
\end{align}
The resultant coefficient vector $\mathbf{c}^{(L+1)}$ represents the encoded state of the time series extended to length $L+1$. Given the coefficient $\mathbf{c}^{(L+1)}$, we can easily construct a time series data of length $L+1$. Let this process be denoted as $\text{hippo}^{-1}$:
\begin{align}
    \text{hippo}_{L+1}^{-1}: \mathbb{R}^N \rightarrow \mathbb{R}^{L+1}, \quad \mathbf{c}^{(L+1)} \mapsto u'_{1:L+1} = \text{hippo}_{L+1}^{-1}(\mathbf{c}^{(L+1)}),
\end{align}
where $u_{1:L}$ and $u'_{1:L+1}$ are different time series. This process effectively extends the original time series by one time step, generating a prediction for the next value in the sequence. By operating within the fixed-dimensional coefficient space $\mathbb{R}^N$, where $N$ is independent of the sequence length $L$, our approach maintains parameter efficiency and scalability. The use of KAN in this context allows for the modeling of complex nonlinear relationships between the coefficients, capturing the underlying dynamics of the time series. 

\subsection{Definition of HiPPO-KAN}
We define the HiPPO-KAN model as a seq2seq mapping that integrates the HiPPO transformations with the KAN mapping. Formally, HiPPO-KAN is defined as
\begin{align}
    \text{HiPPO-KAN} \equiv \text{hippo}_{L+1}^{-1} \circ \text{KAN} \circ \text{hippo}_L.
\end{align}
This composite mapping takes the original time series $\{u_t\}_{t=1}^L$ as input and produces an extended time series $\{u_t\}_{t=1}^{L+1}$ as output:
\begin{align}
    \text{HiPPO-KAN}: \mathbb{R}^L \rightarrow \mathbb{R}^{L+1}, \quad \{u_t\}_{t=1}^L \mapsto \{u'_t\}_{t=1}^{L+1}.
\end{align}
In other words, HiPPO-KAN maps a time series of length $L$ to a different time series of length $L+1$, effectively predicting the next value in the sequence while retaining the original sequence. By integrating these components, HiPPO-KAN effectively captures long-term dependencies and complex temporal patterns in time-series data. Operating within the coefficient space $\mathbb{R}^N$ ensures that the model remains parameter-efficient and scalable, as the dimensionality $N$ does not depend on the sequence length $L$. 

Following the definition of the HiPPO-KAN model, we derive its explicit output formulation by integrating the HiPPO transformations with the KAN mapping. Applying the $\text{hippo}_L$ transformation to the input time series $\{u_t\}_{t=1}^L$, the function $f(s)$ can be approximately represented in terms of orthogonal basis functions:
\begin{align}
    f(s) \approx \sum_{n=1}^N c_n \, p_n(L, s),
\end{align}
where $c_n \in \mathbb{R}$ are the coefficients, and $p_n(L, s)$ are the HiPPO basis functions evaluated at time $L$ for all $s \leq L$. 

Utilizing the KAN mapping, we update the coefficients to incorporate the system dynamics:
\begin{align}
    c'_n = \sum_{m=1}^N \Phi_{nm}(c_m),
\end{align}
where $\Phi_{nm}$ are the elements of the KAN matrix $\Phi \in \mathbb{R}^{N \times N}$. We defined $\text{hippo}^{-1}$ as
\begin{align}
    u'_{1:L+1} = \sum_{n=1}^N \big(c'_n + B u_L\big)\, p_n(L+1, s) = \sum_{n=1}^N \left(\sum_{m=1}^N \Phi_{nm}(c_m) + B u_L \right) p_n(L+1, s),
\end{align}
where $B \in \mathbb{R}^N$ is learnable parameters. This is analogous to the $\mathbf{B} u(t)$ term in the state-space model's state equation. Evaluating at $s = L + 1$, the final output for the next time step is:
\begin{align}
    u'_{L+1} = \sum_{n=1}^N \left(\sum_{m=1}^N \Phi_{nm}(c_m) + B u_L \right) p_n(L+1, L+1).
\end{align}
In the case of Leg-S, from the definition of basis, we have $p_n(L+1, L+1) = \sqrt{2n+1}$ \cite{hippo, hippo-train}. Hence, we obtain
\begin{align}
    u'_{L+1} = \sum_{n=1}^N \sqrt{2n+1} \left(\sum_{m=1}^N \Phi_{nm}(c_m) + B u_L\right).
\end{align}

This methodology resembles an auto-encoder architecture, where the encoder (HiPPO transformation) compresses the input time series into a latent coefficient vector $\mathbf{c}^{(L)}$, the dynamics of which is modelled by KAN layers in our HiPPO-KAN model. The decoder (inverse HiPPO transformation) reconstructs the extended time series from $\mathbf{c}^{(L+1)}$. The fixed-dimensional latent space acts as a bottleneck, promoting efficient learning. 

\subsection{Method}
\subsubsection{Task Definition}
In this study, we address the problem of time series forecasting in the context of cryptocurrency markets, specifically focusing on the BTC-USDT trading pair. The objective is to predict the next price point given a historical sequence of observed prices. Formally, let $\{u_t\}_{t=1}^L$ denote a univariate time series representing the BTC-USDT prices at discrete time steps $t=1,2,\ldots, L$, where $L$ is the window size. The forecasting task aims to estimate the subsequent value $u_{L+1}$ based on the given window of past observations.

Mathematically, the prediction function can be expressed as:
\begin{align}
    \hat{u}_{L+1} = f(u_1,u_2,\ldots,u_L),
\end{align}
where $f: \mathbb{R}^L \rightarrow \mathbb{R}$ is a mapping from the past $L$ observations to the predicted next value $\hat{u}_{L+1}$.

The challenge inherent in this task include:
\begin{itemize}
    \item \textbf{Non-Stationary } Cryptocurrency prices exhibit high volatility and non-stationary behavior, making it difficult to model underlying patterns using traditional statistical methods.
    \item \textbf{Long-Term Dependencies } Capturing long-term dependencies is essential, as market trends and cycles can influence future prices over extended periods.
    \item \textbf{Computational Efficiency } Handling long sequences efficiently without a proportional increase in computational complexity or model parameters is critical for scalability.
\end{itemize}
Our approach utilize the HiPPO-KAN model to effectively tackle these challenges by encoding the input time series into a fixed-dimensional coefficient vector using the HiPPO transformation. This allows the model to process long sequence while maintaining a constant parameter count, facilitating efficient learning and improved predictive accuracy.

\subsubsection{Data Normalization}
We evaluated the performance of HiPPO-KAN using the BTC-USDT 1-minute futures data from January 1st to Feburuary 1st, which consists of univariate time series data. Prior to training, we normalized the raw time series data using the formula $(u_t - \mu) / \mu$, where $\mu$ denotes the mean value of the data within each window. This normalization serves several critical purposes in the context of time series modeling. Firstly, it centers the data around zero, which helps in stabilizing the training process and accelerating convergence by mitigating biases introduced by varying data scales. Secondly, scaling by the mean adjusts for fluctuations in the magnitude of the data across different windows, ensuring that the model's learning is not skewed by windows with larger absolute values.

By normalizing each window individually, we effectively address the non-stationarity inherent in financial time series data, where statistical properties such as mean and variance can change over time. This window-specific normalization allows the model to focus on learning the underlying patterns and dynamics within each window without being influenced by shifts in the data scale. Consequently, this approach enhances the robustness of the model and improves its ability to generalize across different segments of the time series.

\subsubsection{Loss Function for Model Training}
The training of the HiPPO-KAN model involves optimizing the network parameters to minimize the discrepancy between the predicted values and the actual observed values in the time series data. We employ the Mean Squared Error (MSE) as the loss function, which is a standard choice for regression tasks in time series forecasting due to its sensitivity to large errors. 

The MSE loss function is defined as:
\begin{align}
    \mathcal{L}(\theta) = \frac{1}{D} \sum_{i=1}^D \left(
        u^{(i)}_{L+1} - \hat{u}^{(i)}_{L+1}
    \right)^2,
\end{align}
where $\theta$ represents the model parameters, $D$ is the number of samples in the training set, $u^{(i)}_{L+1}$ is the true next value in the time series for the $i$-th sample, and $\hat{u}^{(i)}_{L+1}$ is the corresponding prediction made by the model. 

Minimizing the MSE loss encourages the model to produce predictions that are, on average, as close as possible to the actual values, with larger errors being penalized more heavily due to the squaring operation. The choice of MSE as the loss function aligns with the evaluation metrics used in our experiments, namely the Mean Squared Error (MSE) and Mean Absolute Error (MAE), facilitating a consistent assessment of the model's performance during training and testing.

\subsubsection{Experimental Results}
The experimental results are presented in Tables \ref{tab:performance_horizon1} to facilitate a clear and concise comparison of model performances. Table \ref{tab:performance_horizon1} summarizes the results for a prediction horizon of 1. Each table includes the model name, window size, network width (architecture), Mean Squared Error (MSE), Mean Absolute Error (MAE), and the number of parameters used in the model.

By organizing the results in tabular form, we provide a straightforward means to compare the effectiveness of HiPPO-KAN against baseline models such as HiPPO-MLP, KAN, LSTM, and RNN across different configurations. This structured presentation highlights the consistency and scalability of HiPPO-KAN, especially in terms of parameter efficiency and predictive accuracy over varying window sizes and prediction horizons. The tables clearly demonstrate that HiPPO-KAN achieves superior performance with fewer parameters, emphasizing the advantages of integrating HiPPO transformations with KAN mappings in time series forecasting tasks.

We present additional experimental results in Appendix~\ref{appendix:additional_results}.
To evaluate the scalability of the HiPPO-KAN model, we test
its performance of HiPPO-KAN on even larger window sizes. Furthermore, we demonstrate that information bottleneck theory can be effectively applied within the HiPPO-KAN framework.

\subsection{Lagging problem}
While the result presented above are impressive, we observed that the model still suffers from the lagging problem when examining the plots of the predictions. The lagging problem refers to the phenomenon where the model's predictions lag behind the actual time series, failing to capture sudden changes promptly \cite{lagging_problem}. This issue is particularly detrimental in time series forecasting, where timely and accurate predictions are crucial.

\begin{table}[H]
\centering
\caption{Performance comparison of models for prediction horizon 1. Best models are highlighted in bold.}
\label{tab:performance_horizon1}
\begin{tabular}{lcccccc}
\hline
\textbf{Model} & \textbf{Window Size} & \textbf{Width} & \textbf{MSE} & \textbf{MAE} & \textbf{Parameters} \\
\hline
HiPPO-KAN & 120  & [16, 16]        & $3.40 \times 10^{-7}$  & $4.14 \times 10^{-4}$  & 4,384 \\
HiPPO-KAN & 500  & [16, 16]        & $3.34 \times 10^{-7}$ & $\mathbf{3.95 \times 10^{-4}}$  & 4,384 \\
HiPPO-KAN & 1200 & [16, 16]        & $\mathbf{3.26 \times 10^{-7}}$ & $4.00 \times 10^{-4}$  & 4,384 \\
HiPPO-MLP & 120  & [32, 64, 64, 32, 32] & $2.33 \times 10^{-6}$ & $1.04 \times 10^{-3}$ & 9,792 \\
HiPPO-MLP & 500  & [32, 64, 64, 32, 32] & $2.68 \times 10^{-5}$                & $3.84 \times 10^{-3}$                 & 9,792 \\
HiPPO-MLP & 1200 & [32, 64, 64, 32, 32] & $5.87 \times 10^{-6}$ & $1.96 \times 10^{-3}$ & 9,792 \\
KAN       & 120  & [120, 1]        & $8.9 \times 10^{-7}$  & $6.82 \times 10^{-4}$  & 1,680 \\
KAN       & 500  & [500, 1]        & $1.66 \times 10^{-6}$ & $9.62 \times 10^{-4}$  & 7,000 \\
KAN       & 1200 & [1200, 1]       & $4.03 \times 10^{-6}$ & $1.56 \times 10^{-3}$  & 16,800 \\
LSTM      & 120  & -               & $4.69 \times 10^{-7}$ & $4.99 \times 10^{-4}$  & 4,513 \\
LSTM      & 500  & -               & $6.50 \times 10^{-7}$ & $6.00 \times 10^{-4}$  & 4,513 \\
LSTM      & 1200 & -               & $9.21 \times 10^{-7}$ & $7.21 \times 10^{-4}$  & 4,513 \\
RNN       & 120  & -               & $1.14 \times 10^{-6}$ & $8.60 \times 10^{-4}$  & 12,673 \\
RNN       & 500  & -               & $1.09 \times 10^{-6}$ & $7.70 \times 10^{-4}$  & 12,673 \\
RNN       & 1200 & -               & $1.18 \times 10^{-6}$ & $7.79 \times 10^{-4}$  & 12,673 \\
\hline
\end{tabular}
\end{table}

\begin{figure}[H]
    \centering
    \includegraphics[width=\textwidth]{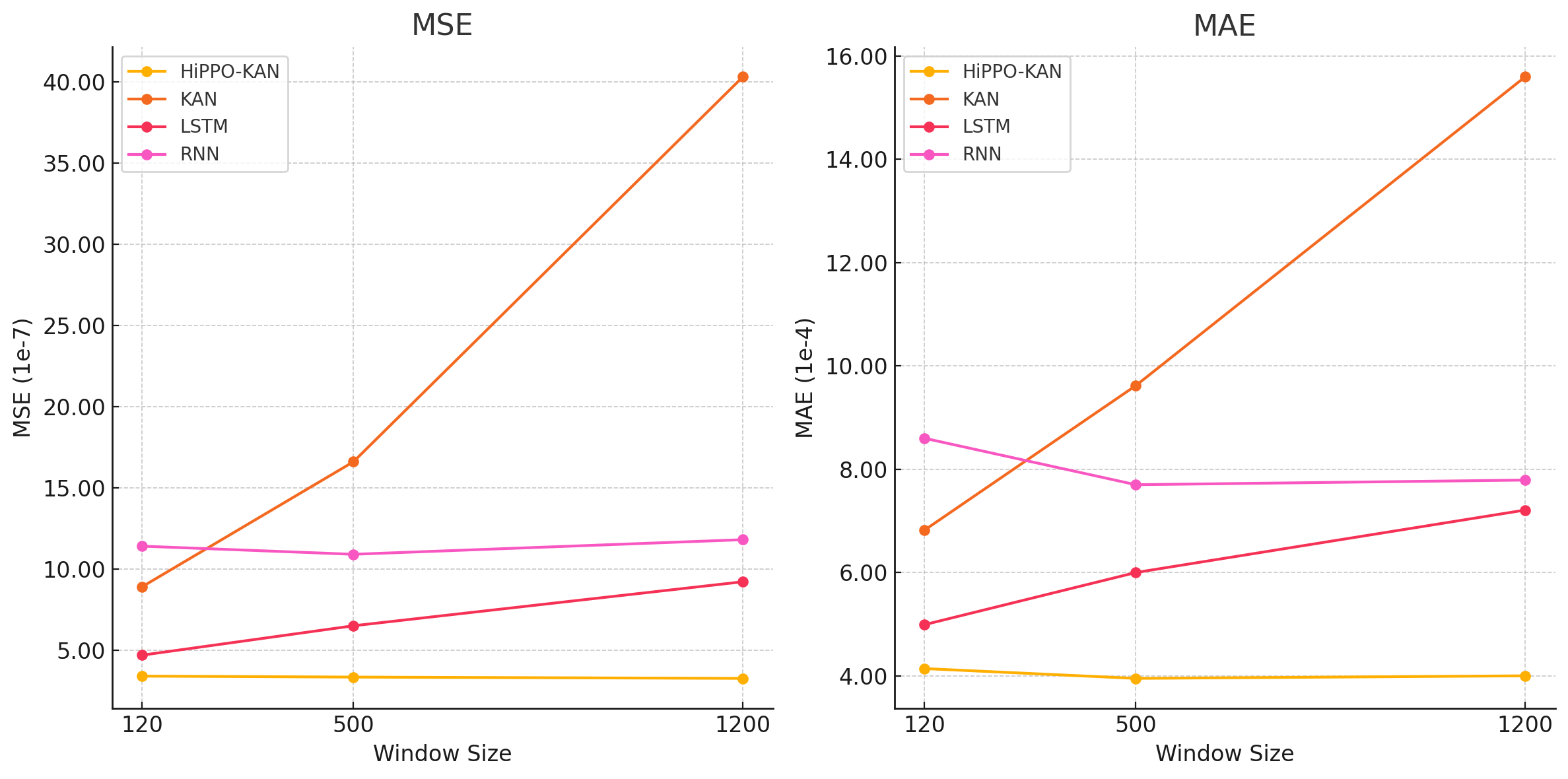}
    \caption{MSE and MAE comparisons for various models (HiPPO-KAN, KAN, LSTM, RNN) using different window sizes (120, 500, 1200). The results show the performance of each model in terms of error metrics as the window size increases.}
    \label{fig:performance comparison}
\end{figure}

To address this issue, we modified the loss function used during training and put $B = 0$. Instead of computing the MSE between the inverse-HiPPO-transformed outputs $\hat{u}_{1:L+1} = \text{hippo}^{-1}_{L+1}\big(\hat{\mathbf{c}}^{(L+1)}\big)$ and the actual time series $u_{1:L+1}$, we computed the MSE directly on the coefficient vectors in the HiPPO domain. Specificaly, the loss function is defined as:

\begin{align}
\mathcal{L}(\theta) = \frac{1}{D} \sum_{i=1}^D \left| \mathbf{c}_{\text{true}}^{(L+1)(i)} - \hat{\mathbf{c}}^{(L+1)(i)} \right|^2
\end{align}

where $\theta$ represents the model parameters, $D$ is the number of samples in the training set, $\mathbf{c}^{(L+1)(i)}_{\text{true}} = \text{hippo}_{L+1}\big( u_{1:L+1}^{(i)} \big)$ is the true coefficient vector obtained by applying the HiPPO transformation to the actual time series, and $\hat{\mathbf{c}}^{(L+1)(i)} = \text{KAN}\big( \mathbf{c}^{(L)(i)} \big)$ is the predicted coefficient vector output by the KAN model. 

By training the model using this modified loss function, we aimed to align the learning process more closely with the underlying representation in the coefficient space, where the HiPPO transformation captures the essential dynamics of the time series. This approach emphasizes learning the progression of the coefficient directly, which may help the model respond more promptly to changes in the input data.

\subsubsection{Interpretation of the Coefficient-Based Loss Function}
\textbf{Representation of Functions in Finite-Dimensional Space } When obtaining the coefficient vector $\mathbf{c}$, it is important to recognize that $\mathbf{c}$ does not represent a single, unique function. Instead, it encapsulates an approximation of the original time series function within a finite-dimensional subspace spanned by the first $N$ basis functions. The approximated function $f(s)$ of a function $f_{\text{true}}$ can be expressed as:
\begin{align}
    f_{\text{true}}(s) = \sum_{i=1}^N c_i p_i(t, s) + \sum_{i=N+1}^\infty c_i p_i(t, s) = f(s) + \sum_{i=N+1}^\infty c_i p_i(t, s)
\end{align}
where $p_i(t,s)$ are the orthogonal basis functions of the HiPPO transformation, and $c_i$ are the corresponding coefficients. The finite sum over $i=1$ to $N$ captures the primary components of the function, while the infinite sum over $i=N+1$ to $\infty$ represents the residual components not captured due to truncation at $N$. This means that $\mathbf{c}$ represents a class of functions sharing the same coefficients for the first $N$ basis functions but potentially differing in higher-order terms. By working with this finite-dimensional approximation, the model focuses on the most significant features of the time series, enabling efficient learning and generalization.

\noindent
\textbf{Impact of Batch Training on Loss Computation } In our training process, we utilize batch training, where the model parameters are updated based on the mean loss computed over a batch of samples. Specifically, the loss function computes the average MSE between the predicted and true coefficient vectors across the batch:
\begin{align}
    \mathcal{L}(\theta) = \frac{1}{D}\sum_{i=1}^D \left|\mathbf{c}_{\text{true}}^{(L+1)(i)} - \hat{\mathbf{c}}^{(L+1)(i)} \right|^2,
\end{align}
where $D$ is the batch size. This approach means that the model learns to minimize the average discrepancy between the predicted and actual coefficients over various time series segments within the batch. 

\noindent
\textbf{Convergence to a Specific Function Through Batch Averaging } By minimizing the average loss across the batch, the model effectively converges towards a specific coefficient vector $\mathbf{c}$ that represents the common underlying dynamics present in the batch samples. This process is akin to converging to a specific function among the possible ones within the function space defined by the finite-dimensional basis. The batch averaging acts as a mechanism to align the model's prediction with the shared features across different time series segments, guiding it towards a consensus representation.

As a result, the model captures the dominant patterns and trends that are consistent across the batch, enhancing its ability to generalize and reducing the likelihood of overfitting to specific instances. The batch mean effectively smooths out idiosyncratic variations in individual samples, promoting the learning of robust features pertinent to the forecasting task. This convergence towards a specific function helps the model to produce more accurate and reliable predictions, particularly when dealing with complex and noisy time series data. 

\noindent
\textbf{Advantages of the Legendre Basis with Exponential Decay Weighting } In the HiPPO transformation, the choice of the Leg-S plays a crucial role in enhancing the model's predictive capabilities. The Leg-S approximation employs a weighting scheme with exponential decay, meaning that the weights assigned to past inputs decrease exponentially over time. This weighting effectively implements a memorization scheme that places more emphasis on the present than on the past. As a result, recent inputs have a stronger influence on the model's state representation than older inputs.

This characteristic leads to more accurate approximations near the final boundary of the time interval, specifically at the prediction point $s = t = L+1$. Since the model assigns greater importance to recent data, the approximated function $f(s)$ closely matches the actual time series values in the neighborhood of the final boundary. Therefore the Legendre basis functions in the Leg-S approximation provide almost equal to the actual values at the final time steps.

\begin{figure}[H]
    \centering
    \includegraphics[width=0.9\textwidth]{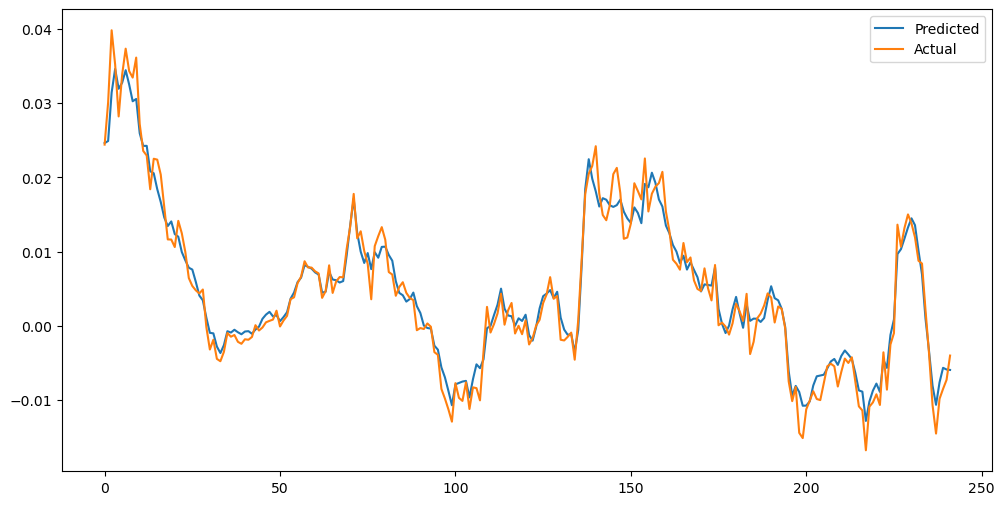}
    \caption{The modified loss function effectively resolves the lagging problem, resulting in predictions that closely follow the actual time series data. This result is based on a randomly selected segment of BTC-USDT 1-minute interval data, using a KAN architecture with a width of [16, 2, 16].}
    \label{fig:hippo loss}
\end{figure}

By leveraging the exponential decay weighting of the Leg-S basis, the model can produce predictions that closely follow the actual data where it matters most—the immediate future. This enhanced accuracy at the prediction boundary is particularly beneficial in time series forecasting applications, where capturing sudden changes and trends promptly is crucial for timely and accurate predictions. The ability to emphasize recent observations allows the HiPPO-KAN model to be more responsive to new information, effectively mitigating issues like the lagging problem and improving overall forecasting performance. As illustrated in Figure~\ref{fig:hippo loss}, the predictions made by the model are now more accurately aligned with the actual time series, effectively capturing sudden changes without delay. Additional results can be found in Appendix~\ref{appendix:additional_results}.

\begin{figure}[H]
    \centering
    \includegraphics[width=0.9\textwidth]{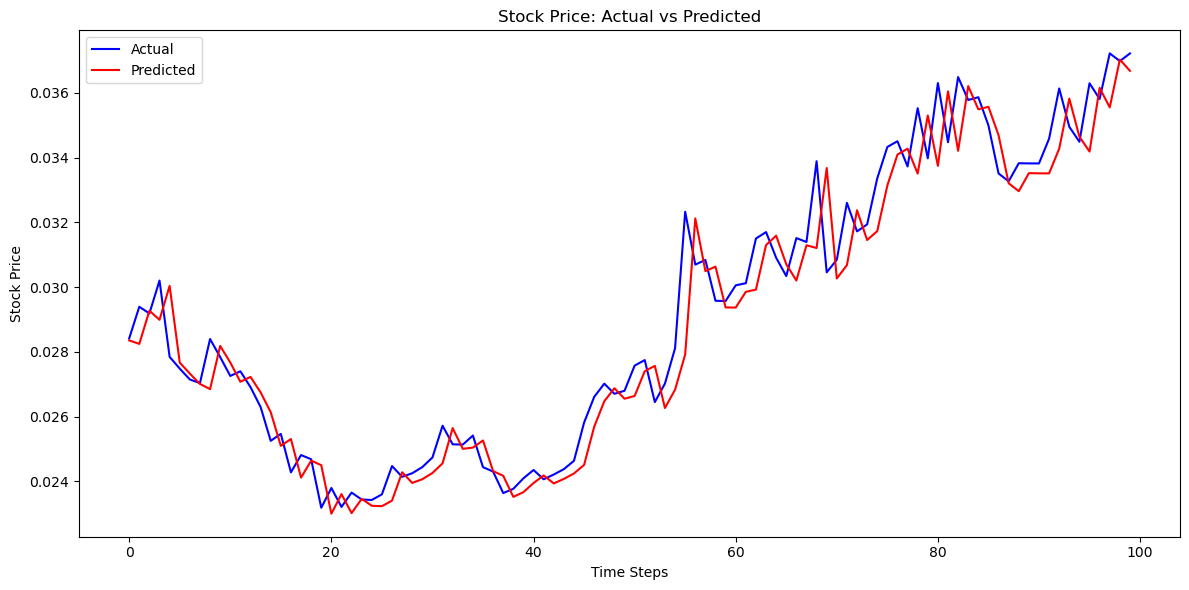}
    \caption{Lagging Effect in KAN Models. These models exhibit a tendency to produce outputs that closely mimic the preceding values, indicating an inability to capture rapid changes in the data effectively.}
    \label{fig:lagging}
\end{figure}

\section{Conclusion}
In this study, we introduced HiPPO-KAN, a novel model that integrates the HiPPO framework with the KAN model to enhance time series forecasting. By encoding time series data into a fixed-dimensional coefficient vector using the HiPPO transformation, and then modeling the progression of these coefficients with KAN, HiPPO-KAN efficiently performed time-series prediction task. 

Our experimental results, as presented in Table \ref{tab:performance_horizon1}, demonstrate that HiPPO-KAN consistently outperforms traditional KAN and other baseline models such as HiPPO-MLP, LSTM, and RNN across various window sizes and prediction horizons. Notably, HiPPO-KAN maintains a constant parameter count regardless of sequence length, highlighting its parameter efficiency and scalability. For example, at a window size of 1,200 and a prediction horizon of 1, HiPPO-KAN achieved an MSE of $3.26 \times 10^{-7}$ and an MAE of $4.00 \times 10^{-4}$, compared to KAN's MSE of $4.03 \times 10^{-6}$ and MAE of $1.56 \times 10^{-3}$, with fewer parameters.

The integration of HiPPO theory into the KAN framework provides a powerful approach for handling long sequences without increasing the model size. By operating within a fixed-dimensional latent space, HiPPO-KAN not only improves predictive accuracy but also offers better interpretability of the model's internal workings. The use of KAN allows for modeling complex nonlinear relationships between the HiPPO coefficients, capturing the underlying dynamics of the time series more effectively than traditional methods. These promising results position HiPPO-KAN as a significant advancement in time-series forecasting, offering a scalable and efficient solution that could potentially revolutionize applications across various domains, from financial modeling to climate prediction.

Additionally, we addressed the lagging problem commonly encountered in time series forecasting models. By modifying the loss function to compute the MSE directly on the coefficient vectors in the HiPPO domain, we significantly improved the model's ability to capture sudden changes in the data without delay. This adjustment aligns the learning process more closely with the underlying dynamics of the time series, allowing HiPPO-KAN to produce predictions that closely follow the actual data, as illustrated in Figure~\ref{fig:hippo loss}.

\subsection{Future Work}
\textbf{Integration with Graph Neural Networks for Multivariate Time Series }\\
To extend HiPPO-KAN to handle multivariate time series data, we propose integrating it with Graph Neural Networks (GNNs) \cite{geometric_dl}. In this framework, each variable or time series in the multivariate dataset is represented as a node within a graph structure. At each node, the HiPPO transformation encodes the local time series data into a fixed-dimensional coefficient vector, analogous to a gauge vector in physics.

These gauge-like vectors serve as localized representations of the temporal dynamics at each node. The edges of the graph define the interactions between nodes, capturing the dependencies and relationships among different variables in the dataset. By modeling these interactions, we can define functions that operate on pairs or groups of coefficient vectors, effectively allowing information to flow across the graph and capturing the multivariate dependencies.

This integration leverages the strength of HiPPO-KAN in modeling individual time series efficiently while utilizing the relational modeling capabilities of GNNs to handle the interconnectedness of multivariate data. Future work could focus on developing this combined HiPPO-KAN-GNN architecture, investigating how the interactions between nodes can be effectively modeled, and exploring the impact on forecasting accuracy and interpretability. This approach has the potential to address complex systems where variables are interdependent, such as in financial markets, climate modeling, and social network analysis.


\bibliographystyle{unsrt}
\bibliography{main}

\newpage
\appendix

\section*{APPENDIX}
\section{Additional Experimental Results}
\label{appendix:additional_results}

To further elucidate the scalability and robustness of the HiPPO-KAN model, we conducted a series of experiments with larger window sizes of 2500, 3000, 3500, and 4000. The results are summarized in Table~\ref{tab:additional_performance}. Our findings indicate that the model's accuracy experiences only marginal degradation as the window size increases up to 4000. Notably, while the window size expands by a factor of 33, the MSE loss of the model increases by a mere factor of approximately 1.3, with model parameters held constant. This small increase in error relative to the substantial increase in window size demonstrates the exceptional scalability and computational efficiency of the HiPPO-KAN model. These results not only demonstrate the model's resilience to increased input complexity but also underscore its potential for application in scenarios demanding the processing of extensive temporal sequences without significant compromise in performance.

\begin{table}[H]
\centering
\caption{Performance of HiPPO-KAN on larger window sizes. Best models are highlighted in bold.}
\label{tab:additional_performance}
\begin{tabular}{lccccc}
\hline
\textbf{Model} & \textbf{Window Size} & \textbf{Width} & \textbf{MSE} & \textbf{MAE} & \textbf{Parameters} \\
\hline
HiPPO-KAN & 2500 & [16, 16] & $3.33 \times 10^{-7}$ & $4.13 \times 10^{-4}$ & 4384\\
HiPPO-KAN & 3000 & [16, 16] & $3.68 \times 10^{-7}$ & $4.41 \times 10^{-4}$ & 4384\\
HiPPO-KAN & 3500 & [16, 16] & $4.01 \times 10^{-7}$ & $4.66 \times 10^{-4}$ & 4384\\
HiPPO-KAN & 4000 & [16, 16] & $4.38 \times 10^{-7}$ & $4.89 \times 10^{-4}$ & 4384\\
HiPPO-KAN & 2500 & [16, 2, 16] & $\mathbf{3.10 \times 10^{-7}}$ & $\mathbf{3.90 \times 10^{-4}}$ & 1344\\
HiPPO-KAN & 3000 & [16, 2, 16] & $3.29 \times 10^{-7}$ & $4.05 \times 10^{-4}$ & 1344\\
HiPPO-KAN & 3500 & [16, 2, 16] & $3.46 \times 10^{-7}$ & $4.19 \times 10^{-4}$ & 1344\\
HiPPO-KAN & 4000 & [16, 2, 16] & $3.96 \times 10^{-7}$ & $4.50 \times 10^{-4}$ & 1344\\
HiPPO-KAN & 2500 & [16, 4, 16] & $3.13 \times 10^{-7}$ & $3.94 \times 10^{-4}$ & 2400\\
HiPPO-KAN & 3000 & [16, 4, 16] & $3.29 \times 10^{-7}$ & $4.05 \times 10^{-4}$ & 2400\\
HiPPO-KAN & 3500 & [16, 4, 16] & $4.03 \times 10^{-7}$ & $4.66 \times 10^{-4}$ & 2400\\
HiPPO-KAN & 4000 & [16, 4, 16] & $3.84 \times 10^{-7}$ & $4.48 \times 10^{-4}$ & 2400\\
\hline
\end{tabular}
\end{table}

\begin{figure}[H]
    \centering
    \includegraphics[width=\textwidth]{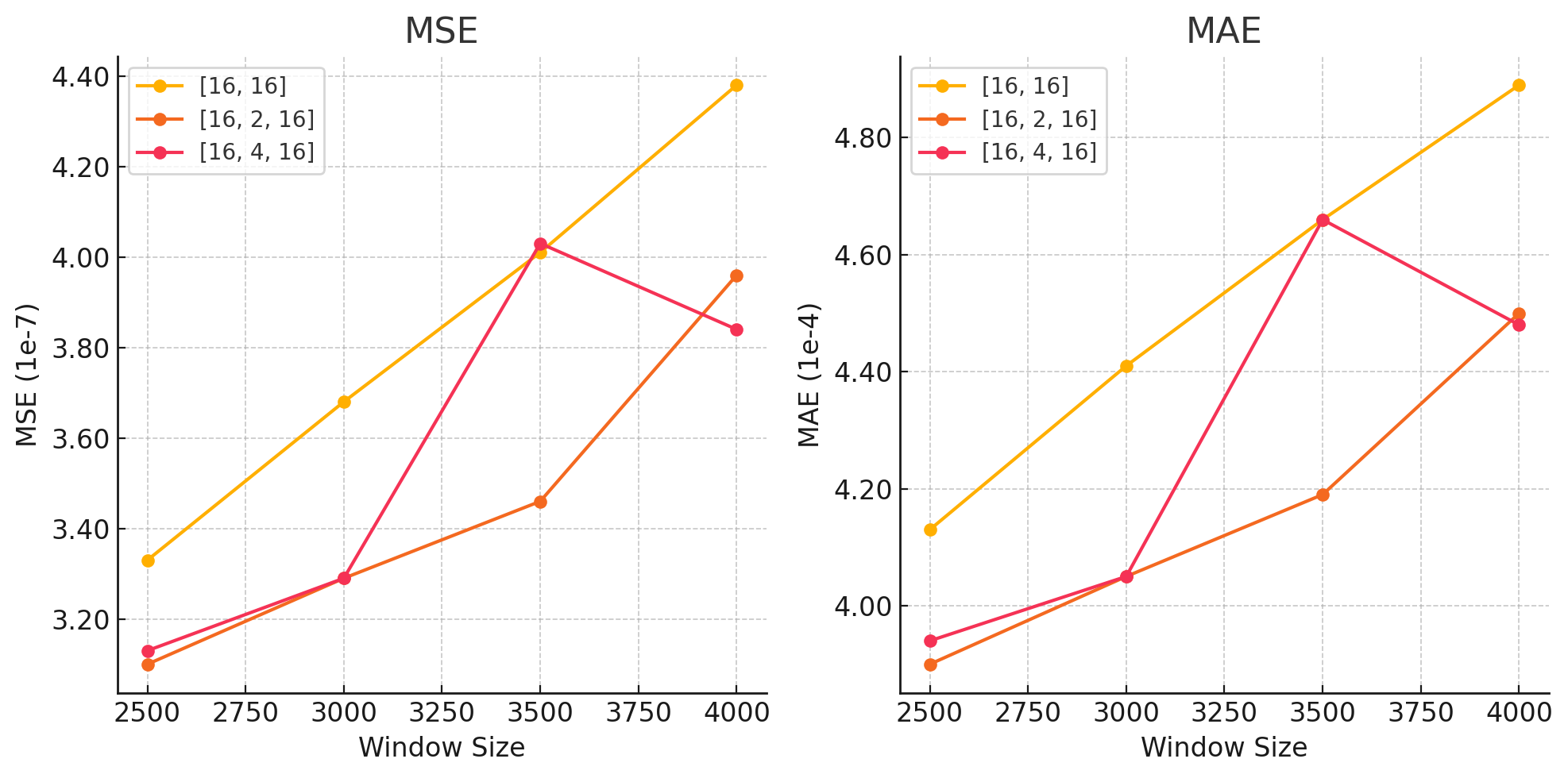}
    \caption{Performance of HiPPO-KAN on larger window sizes, illustrating the impact of the bottleneck layer on model efficiency. Configurations with a bottleneck achieve lower Mean Squared Error (MSE) and Mean Absolute Error (MAE) compared to models without a bottleneck despite having fewer parameters.}
    \label{fig:performance comparison2}
\end{figure}

In addition, we conducted experiments with HiPPO-KAN models incorporating a bottleneck layer within their network architecture. Intriguingly, as demonstrated in Table~\ref{tab:additional_performance} and Fig.~\ref{fig:performance comparison2}, the HiPPO-KAN model featuring a bottleneck layer exhibited better performance compared to its counterpart without such a layer, despite having fewer parameters. This seemingly counterintuitive outcome can be elucidated through the lens of information bottleneck theory \cite{info_bottleneck, info_bottleneck2}. This theoretical framework posits that models can derive benefits from compressing input information, thereby distilling the most salient features pertinent to the prediction task. The enhanced performance of the bottleneck model aligns with this principle, suggesting that the constrained representation enforced by the bottleneck layer facilitates more effective feature extraction and, consequently, improved predictive capability.

\end{document}